\documentclass{article}

\usepackage[final]{neurips_2020}

\usepackage[utf8]{inputenc} 
\usepackage[T1]{fontenc}    
\usepackage[usenames, dvipsnames]{color} 
\usepackage{enumerate, amsmath, amsthm, amssymb,  algpseudocode, pifont, fullpage, csquotes, dashrule, tikz, bbm, booktabs, bm, verbatim, url}
\usepackage[framemethod=TikZ]{mdframed}
\usepackage{natbib}
\usepackage[ruled,vlined]{algorithm2e}
\usepackage[colorlinks]{hyperref}
\hypersetup{
     colorlinks = true,
     linkcolor = blue,
     anchorcolor = blue,
     citecolor = red,
     filecolor = blue,
     urlcolor = blue
     }
\usepackage{caption}
\usepackage{subcaption}

\definecolor{dblue}{RGB}{98, 140, 190}
\definecolor{dlblue}{RGB}{216, 235, 255}
\definecolor{dgreen}{RGB}{124, 155, 127}
\definecolor{dpink}{RGB}{207, 166, 208}
\definecolor{dyellow}{RGB}{255, 248, 199}
\definecolor{dgray}{RGB}{46, 49, 49}

\newcommand{\durl}[1]{\textcolor{dblue}{\underline{\url{#1}}}}

\newcommand{\mc}[1]{\mathcal{#1}}
\newcommand{\mf}[1]{\mathfrak{#1}}

\newcommand{\bE}{\mathbb{E}}
\newcommand{\bV}{\mathbb{V}}
\newcommand{\bR}{\mathbb{R}}

\newcommand{\bN}{\mathbb{N}}
\newcommand{\mbf}[1]{\mathbf{#1}}
\newcommand{\rank}{\text{rank}}
\newcommand{\kl}[2]{D_{\text{KL}}(#1 || #2)}

\DeclareMathOperator*{\argmax}{arg\,max}

\newmdenv[
  topline=false,
  bottomline=false,
  rightline = false,
  leftmargin=10pt,
  rightmargin=0pt,
  innertopmargin=0pt,
  innerbottommargin=0pt
]{innerproof}

\newcounter{DaveDefCounter}
\newcommand{\ddef}[2]
{
\begin{mdframed}[roundcorner=1pt, backgroundcolor=white]
\vspace{1mm}
{\bf Definition \theDaveDefCounter} (#1): {\it #2}
\stepcounter{DaveDefCounter}
\end{mdframed}
}

\newtheorem{assumption}{Assumption}

\newif\ifsubmit
\submitfalse
\ifsubmit
\newcommand{\dnote}[1]{}
\newcommand{\bnote}[1]{}
\else
\newcommand{\dnote}[1]{\textcolor{blue}{Dilip: #1}}
\newcommand{\bnote}[1]{\textcolor{orange}{Ben: #1}}
\fi

\newcommand{\name}{PS2}

\title{Randomized Value Functions via \\ Posterior State-Abstraction Sampling}

\author{%
  Dilip Arumugam\\
  Department of Computer Science\\
  Stanford University\\
  \texttt{dilip@cs.stanford.edu}\\
  \And
  Benjamin Van Roy \\
  Department of Electrical Engineering \\
  Department of Management Science \& Engineering\\
  Stanford University\\
  \texttt{bvr@stanford.edu} \\
}

\begin{document}

\maketitle

\begin{abstract}
    State abstraction has been an essential tool for dramatically improving the sample efficiency of reinforcement-learning algorithms. Indeed, by exposing and accentuating various types of latent structure within the environment, different classes of state abstraction have enabled improved theoretical guarantees and empirical performance. When dealing with state abstractions that capture structure in the value function, however, a standard assumption is that the true abstraction has been supplied or unrealistically computed a priori, leaving open the question of how to efficiently uncover such latent structure while jointly seeking out optimal behavior. Taking inspiration from the bandit literature, we propose that an agent seeking out latent task structure must explicitly represent and maintain its uncertainty over that structure as part of its overall uncertainty about the environment. We introduce a practical algorithm for doing this using two posterior distributions over state abstractions and abstract-state values. In empirically validating our approach, we find that substantial performance gains lie in the multi-task setting where tasks share a common, low-dimensional representation.
\end{abstract}

\section{Introduction}

The complexity of the state space within an environment has profound impact on a decision-making agent's capacity for sample-efficient reinforcement learning~\citep{kearns2002near,kakade2003sample,strehl2009reinforcement,auer2009near}. Oftentimes, however, this complexity (as measured by the cardinality of a finite set of states or the dimensionality of real-valued observations/features) is an exaggeration of the amount of information actually needed to make optimal (or even near-optimal) decisions. And yet, many reinforcement-learning algorithms make no concerted effort to fully exploit this structure so as to accelerate learning of the optimal policy or value function~\citep{watkins1992q,williams1992simple,sutton1988learning,sutton2000policy,mnih2015human}.

In contrast, there exist several works which closely examine state abstractions from a theoretical or empirical perspective, highlighting the advantages of acknowledging and utilizing latent problem structure~\citep{bertsekas1988adaptive,dean1997model,ferns2004metrics,jong2005state,li2006towards,van2006performance,ferns2012methods,jiang2015abstraction,abel2016near,abel2019state,dong2019provably,du2019provably,misra2019kinematic,russo2020approximation}. While important for gaining clarity and an appreciation for what state abstraction brings to the reinforcement-learning problem~\citep{lake2017building,konidaris2019necessity}, some of these works can be unrealistic in their acquisition of the very object being studied. Specifically, those approaches focusing on abstractions that capture structure within the optimal value function often assume that an ideal state abstraction has been provided by an oracle~\citep{li2006towards,abel2016near,dong2019provably}; while this may be a suitable provision for the purpose of theoretical analysis, this engenders a two-step procedure in practice whereby the optimal value function is computed exactly and then used to compute the corresponding state abstraction~\citep{abel2016near,abel2018state}. These approaches beg a natural question: \textit{can an agent learn a value-based state abstraction while simultaneously learning how to act optimally within an environment?} Incorporating this additional abstraction criterion, beyond the standard RL objective of learning an optimal policy, amounts to inserting a form of inductive bias within the learning algorithm. Thus, a natural follow-up question arises: \textit{when is it advantageous to learn a state abstraction alongside learning optimal behavior?}

In this work, we offer an affirmative answer to the first question and provide an empirically-supported hypothesis to the second. In addressing the former, we build upon an existing line of work that grounds (both theoretically and empirically) the efficient learning of optimal behavior through an agent's representation of epistemic uncertainty over its environment~\citep{russo2016information,osband2016deep,russo2018learning,o2018uncertainty,osband2019deep}. Informally, it is the agent's resolution of this uncertainty that naturally fosters deep exploration and provably-efficient learning. Adopting this perspective, we assert that an agent must maintain an explicit belief over the state abstraction that underlies the environment. This belief can then be coupled with a conditional posterior over abstract-state values to render the agent's full uncertainty over the optimal value function. As for when it is advantageous to incorporate this factored posterior, we empirically demonstrate considerable performance gains when an agent engages with multiple tasks in the same environment, all supported by a single, underlying state abstraction.

The paper proceeds as follows: in Section \ref{sec:rank}, we formulate the problem and clarify our setting where the optimal value function admits a natural state abstraction, which we formalize as exhibiting low-rank structure. We then introduce our approach, Posterior State-abstraction Sampling (\name), in Section \ref{sec:appr}. We conclude with illustrative experiments in contextual-bandit problems, exploring both the single-task and multi-task learning settings (Sections \ref{sec:exps} and \ref{sec:conc}). Due to space constraints, we defer a full presentation of background material and related work to the appendix.

\section{Problem Formulation}
\label{sec:rank}

\subsection{Reinforcement Learning}

We formulate an individual task as a finite-horizon, episodic Markov Decision Process (MDP)~\citep{bellman1957markovian,Puterman94} defined by $\mc{M} = \langle \mc{S}, \mc{A}, \mc{R}, \mc{T}, \gamma, H \rangle$ where $\mc{S}$ denotes the state space, $\mc{A}$ is the finite action set, $\mc{R}_h:\mc{S} \times \mc{A} \mapsto \bR$ is a (deterministic) reward function for timestep $h$, $\mc{T}:\mc{S} \times \mc{A} \mapsto \Delta(\mc{S})$ is the transition function, $\gamma \in [0,1)$ is the discount factor, and $H \in \bN$ is the horizon or episode duration. We use $|\mc{A}| = A$ and, when $\mc{S}$ is finite, $|\mc{S}| = S$ to denote the respective sizes of the action space and state space.

Learning proceeds in $T$ stages or episodes where, at each timestep of the current episode $h \in [H]$, the agent observes the current state $s_h$ and selects an action $a_h$ according to its current policy $\pi_h:\mc{S} \mapsto \Delta(\mc{A})$. We assume that $\mc{M}$ has a fixed initial state distribution $\rho \in \Delta(\mc{S})$ such that $s_1 \sim \rho(\cdot)$ is observed at the start of each episode. The objective for the agent is to synthesize a policy so as to maximize expected return $\bE[\sum\limits_{h=1}^H \gamma^{h-1} \mc{R}_h(s_h,a_h)]$. The value function of a policy $\pi$ denotes the expected future discounted return by following the policy from a given state $s$, $V_h^\pi(s_h) = \bE[\sum\limits_{h'=0}^{H-h} \gamma^{h'}\mc{R}_h(s_{h'}, a_{h'}) | s_{h'} = s_h]$, where the expectation is taken with respect to the stochasticity in the environment transition dynamics and policy.Similarly, we use the action-value function $Q^\pi(s,a)$ to define the expected future discounted return from being in a state $s$, taking action $a$, and following policy $\pi$ thereafter,$Q_h^\pi(s, a) = \mc{R}_h(s,a) + \gamma \bE_{s' \sim \mc{T}(\cdot | s, a)}[V_{h+1}^\pi(s')]$. Denoting the policy class containing all stationary, stochastic policies as $\Pi = \{\pi | \pi:\mc{S} \mapsto \Delta(\mc{A}) \}$, we may define the optimal policy $\pi^\star = \sup\limits_{\pi \in \Pi^H} V_1^\pi(s)$, where $\Pi^H$ denotes the class of non-stationary policies, whose value functions are given by the Bellman optimality equations: $V_h^\star(s) = \max\limits_{a \in \mc{A}} Q_h^\star(s,a)$ and $Q_h^\star(s, a) = \mc{R}_h(s,a) + \gamma \bE_{s' \sim \mc{T}(\cdot | s, a)}[V_{h+1}^\star(s')]$, where $V_{H+1}^\star(s) = 0$. Concretely, our goal is to leverage experience sampled in each episode to learn $Q^\star$, yielding the optimal policy $\pi_h^\star(s) = \argmax\limits_{a \in \mc{A}} Q_h^\star(s,a)$~\citep{sutton1998introduction,watkins1992q}.

\subsection{State Abstractions}

A well-studied tool for accelerating RL algorithms is the use of state aggregation or state abstraction to reduce the size of the MDP state space~\citep{bertsekas1988adaptive,li2006towards,van2006performance}. Indeed, given the dependence on $|\mc{S}|$ that appears in numerous sample complexity results for RL~\citep{kakade2003sample,strehl2009reinforcement}, it seems only natural that a reduction in the overall number of states under consideration can dramatically alleviate the burdens of learning an optimal policy.

As outlined in \citet{li2006towards}, several classes of state abstractions exist, each elucidating structure contained in different components of the MDP. A choice of state abstraction from one of these classes characterizes a particular function $\phi: \mc{S} \mapsto \mc{S}_\phi$ mapping original or ground states of the MDP into an aggregate or abstract state space $\mc{S}_\phi$. Naturally, the abstract state space is taken to be smaller, in some sense, than the original (for instance, $|\mc{S}_\phi| < |\mc{S}|$) such that $\phi$ defines a (lossy) compression of the original state space~\citep{abel2019state}. While state abstraction constitutes a general mechanism for specifying latent task structure, several works often make generous assumptions concerning the provision or acquisition of $\phi$ before highlighting the benefits of $\phi$ in RL. In contrast, this work weakens those assumptions specifically by learning $\phi$ concurrently with $Q^\star$.

Various prior works have focused on the capacity of state abstraction to enable provably-efficient RL algorithms that scale to tasks with high-dimensional observations~\citep{du2019provably,misra2019kinematic,agarwal2020flambe}. These approaches build state abstractions based on latent structure in the transition dynamics; by operating in the reward-free setting~\citep{hazan2019provably,jin2020reward}, which lacks a specific reward function, these algorithms employ various mechanisms to facilitate thorough exploration of the environment, yielding a strong approximation of the transition model. This style of approach seems like a natural ``path of least resistance'' in so far as each step within the environment yields a ground-truth signal that can be aimed at further distilling the true, underlying abstraction.  In contrast, a state abstraction based on the similarity of $Q^\star$-values~\citep{li2006towards,abel2016near,abel2019state,dong2019provably} inherently draws upon knowledge of $Q^\star$ which, if readily accessible to the agent, would imply knowledge of the optimal policy itself. Nevertheless, in this paper, we explicitly direct our focus to these $Q^\star$-based state abstractions and avert the apparent ``chicken-or-egg'' problem through the agent's own epistemic uncertainty about the environment.

\subsection{Low-Rank Value Functions}

We formalize the type of $Q^\star$-based state abstraction studied in this work under the following definition of a low-rank value function:
\ddef{Low-Rank $Q^\star$-function}{
    The optimal action-value function $Q^\star$ of an MDP $\mc{M}$ is characterized as low-rank if there exists two functions, $\phi_h^\star:\mc{S} \mapsto \bR^M$ and $\psi_h^\star:\mc{A} \mapsto \bR^M$, such that $\forall s \in \mc{S}, a \in \mc{A}, h \in [H]$:
    \begin{align*}
        Q_h^\star(s,a) &= \langle \phi_h(s), \psi_h(a) \rangle
    \end{align*}
    where $\langle u,v \rangle$ denotes the inner product between vectors $u,v$. We interpret the latent dimension $M \in \bN$ as the number of abstract states.
    \label{def:low_rank}
}

\begin{assumption}
Throughout this work, we will assume that $\mc{M}$ admits a low-rank $Q^\star$-function with rank $M$.
\label{asm:low_rank}
\end{assumption}

To clarify this structural assumption, consider the case where $\mc{M}$ has a finite state-action space. Dropping the timestep subscript for clarity, note that $Q^\star$ may then be compactly represented as a matrix $\mbf{Q}^\star \in \bR^{S \times A}$ where $\mbf{Q}_{ij}^\star = Q^\star(s_i, a_j)$, for some arbitrary indexing of states and actions. 
Assumption \ref{asm:low_rank} then follows as $\rank(\mbf{Q}^\star) = M$. In words, we assume that the matrix of optimal action-values admits a low-rank factorization $\mbf{Q}^\star = \mbf{\Phi}\mbf{\Psi}^T$, with  $\mbf{\Phi} \in \bR^{S \times M}$, $\mbf{\Psi} \in \bR^{A \times M}$. Under this view, we can interpret learning a low-rank decomposition of $\mbf{Q}^\star$ as constructing predictive features in a $M$-dimensional subspace that enable accurate estimation of $Q^\star$ for each state-action pair~\citep{tsitsiklis1996feature}. 

We conclude this section by briefly making explicit some connections between low-rank value functions and the $Q^\star$-similarity state abstractions employed in prior work. Specifically, notice that the exact $Q^\star$-irrelevance abstraction of \citet{li2006towards} corresponds to an abstraction function $\phi^\star:\mc{S} \mapsto \{0,1\}^M$ where $\forall s \in \mc{S}, \sum\limits_{i=1}^M \phi(s)_i = 1$; in words, $\phi^\star$ maps each state to a one-hot vector thereby guaranteeing that $\phi(s_1) = \phi(s_2) \iff Q^\star(s_1, a) = Q^\star(s_2,a), \forall s_1,s_2 \in \mc{S}, a \in \mc{A}$.  An approximate version of this abstraction, as studied in \citep{abel2016near,dong2019provably}, can be achieved by maintaining the same conditions on $\phi^\star$ and weakening the definition of a low-rank value function to $|Q^\star(s,a) - \langle \phi^\star(s) \psi^\star(a) \rangle | \leq \frac{\epsilon}{2}$, for some constant $\epsilon$. Finally, the soft state aggregations of \citet{singh1995reinforcement} impose the alternative condition $\phi^\star:\mc{S} \mapsto \Delta^{M-1}$, mapping individual states to the $(M-1)$-dimensional probability simplex.

\section{Approach}
\label{sec:appr}

\subsection{Hypermodels}

When attempting to represent epistemic uncertainty, scalability becomes an immediate challenge. When using neural networks, the common practice has been to employ finite ensembles~\citep{osband2016deep,lu2017ensemble} which maintain several copies of network weights. A sample from this posterior of $K$ ensemble members is acquired by first sampling $z \sim \text{Uniform}(K)$ and then running the ensemble member indexed by $z$.

Since the efficiency of ensemble sampling scales poorly with $K$, \citet{dwaracherla2020hypermodels} introduce hypermodels as a scalable alternative to representing epistemic uncertainty. In short, hypermodels rely on a space of indexes $\mc{Z}$ as well as a corresponding reference distribution $p_z \in \Delta(\mc{Z})$. For a given choice of base model $f_\theta:\mc{X} \mapsto \mc{Y}$ parameterized by $\theta \in \Theta$, a hypermodel with parameters $\nu$, $\mf{H}_\nu:\mc{Z} \mapsto \Theta$ maps a single index to a particular instantiation of base model. Thus, by sampling an index $z \sim p_z(\cdot)$, the function $f_{\mf{H}_\nu(z)}$ represents a sample from an approximate posterior. Given a dataset $\mc{D}$, the loss function $\mc{L}(\nu,\mc{D})$ to optimize hypermodel parameters $\nu$ will vary depending on the base model and task. We defer the definition of $\mc{L}(\nu,\mc{D})$ to the next section.

\subsection{Approximate Posterior over State Abstractions}

Our core contribution is an approach for jointly learning a state abstraction $\phi$  without prior knowledge of $Q^\star$. To do this, we leverage our assumption of $Q^\star$ as being low rank and we explicitly maintain two separate hypermodels $\mf{H}^\phi_\nu,\mf{H}^\psi_\nu$ as approximate posterior distributions over $\phi^\star$ and $\psi^\star$ respectively. It is important to note that while $\mf{H}^\phi_\nu:\mc{Z} \mapsto \Phi$ is a standard hypermodel~\citep{dwaracherla2020hypermodels} mapping indices to instances of $\phi$, $\mf{H}^\psi_\nu:\mc{Z} \times \Phi \mapsto \Psi$ is a conditional hypermodel over possible functions $\psi$. The intuition here is that once an agent samples from its posterior beliefs over $\phi^\star$, the corresponding sample from its beliefs over abstract-state values must be conditioned on the particular sample $\phi \sim \mf{H}^\phi_\nu(\cdot)$. More succinctly, an agent's posterior beliefs over the optimal value function is obtained by first sampling $z \sim p_z(\cdot)$ and then $\hat{Q}^\star \sim \langle \mf{H}^\phi_\nu(z),\mf{H}^\psi_\nu(z,\mf{H}^\phi_\nu(z))\rangle$.

Recall that we have yet to define the objective function for optimizing the hypermodels to represent an approximate posterior over $Q^\star$. Given a minibatch of past experiences $\tilde{\mc{D}}$, we optimize the following loss function $\mc{L}(\nu,\tilde{\mc{D}})$:
\begin{align*}
    R(z_\phi, z_\psi, \nu) &\triangleq \lambda||\mf{H}^\phi_\nu(z_\phi) - \mf{H}^\phi_{\nu_0}(z_\phi)||^2_2 + \lambda||\mf{H}^\psi_\nu(z_\psi, \mf{H}^\phi_\nu(z_\phi)) - \mf{H}^\psi_{\nu_0}(z_\psi, \mf{H}^\phi_{\nu_0}(z_\phi))||^2_2 \\
    \hat{Q}^\star_{z_\phi,z_\psi}(s,a) &\triangleq \langle \mf{H}^\phi_\nu(z_\phi)(s),\mf{H}^\psi_\nu(z_\psi,\mf{H}^\phi_\nu(z_\phi))(a)\rangle \\
    \mc{L}(\nu,\tilde{\mc{D}}) = \bE_{\substack{z_\phi \sim p_z(\cdot) \\ z_\psi \sim p_z(\cdot)}}[\frac{1}{m} &\sum\limits_{(s,a,r,s' \eta_\phi, \eta_\psi) \in \tilde{\mc{D}}}(r + \gamma \max\limits_{a' \in \mc{A}}     \hat{Q}^\star_{z_\phi,z_\psi}(s',a') + \eta_\phi^Tz_\phi + \eta_\psi^Tz_\psi - \hat{Q}^\star_{z_\phi,z_\psi}(s,a))^2 +  R(z_\phi, z_\psi, \nu)]
\end{align*}

where $\nu_0$ denotes the initial vector of hypermodel parameters, the $\eta_\phi, \eta_\psi$ terms denote random Gaussian perturbations of the target values, and $\lambda$ is a regularization coefficient. This loss function encapsulates a randomized least-squares value iteration (RLSVI)~\citep{osband2016generalization} approach to maintaining an approximate posterior distribution over $Q^\star$.

\subsection{Posterior State-Abstraction Sampling}

Due to space constraints, we present \name\ as Algorithm \ref{alg:qmf} in the appendix with the explicit choice of hypermodels for representing an agent's beliefs. The algorithm proceeds according to variance-IDS by first drawing $K$ samples from the agent's current posterior beliefs over $Q^\star$ and computing the requisite quantites for variance-IDS, namely the expected squared regret and expected variance for each action. The resulting experience collected from the environment is then accumulated via experience replay~\citep{lin1992self,mnih2015human} for incrementally updating the hypermodels.

For the multi-task learning setting, the algorithm is nearly identical, with the caveat that index samples must be drawn for each hypermodel over abstract-state values (with one hypermodel per task); the hypermodel over state abstractions is shared across all tasks.

\section{Experiments \& Discussion}
\label{sec:exps}

We recall that the primary goal of this paper is twofold, (1) offering \name\ as a practical approach for synthesizing a state abstraction based on structure within $Q^\star$ (as opposed to, for instance, structure in the transition function) and (2) identifying when the pursuit of such latent structure is particularly advantageous. In this section, we outline a recipe for randomly generating contextual bandit problems that allow us to empirically address these goals. 

Concretely, given specific values for the number of states $(S)$, actions $(A)$, and abstract states $(M)$, we generate contextual bandit problems by randomly sampling a reward function with sparse latent structure. The factorization $\mbf{Q}^\star = \mbf{\Phi}\mbf{\Psi}^T$ is formed by first generating $\mbf{\Phi} \in \bR^{S \times M}$ through the random sampling of $S$ one-hot vectors of length $M$. The abstract-state values $\mbf{\Psi} \in \bR^{A \times M}$ are then each drawn uniformly at random from $[0,1]$. For our multi-task experiments, the procedure for sampling $\mbf{\Phi}$ remains unchanged and the procedure for sampling $\mbf{\Psi}_t$ is repeated for each task $t$. We define instantaneous regret as $\Delta_t = |\max\limits_{a^\star} Q^\star(s_t,a^\star) - Q^\star(s_t,a_t)|$ and define cumulative regret over $T$ episodes or time periods as $\sum\limits_{t=1}^T \Delta_t$. For multi-task experiments, instantaneous regret is summed across all tasks. In all experiments we use the Adam optimizer~\citep{kingma2014adam}  with a learning rate of $0.001$, a batch size of $1024$, $128$ index samples per timestep, noise variances of $0.25$, and regularization parameter of $0.001$. All shading in figures denote $95\%$ confidence intervals computed across five random seeds. All hypermodels are parameterized as Gaussian distributions with index samples of appropriate dimension drawn iid from $\mc{N}(0,1)$. For each contextual-bandit problem, we evaluate the following algorithms:
\begin{itemize}
    \item \textbf{PS2-IDS} -- Algorithm \ref{alg:qmf} with two hypermodels over state abstractions and abstract-state values respectively. The former hypermodel is parameterized as a Gaussian distribution with diagonal covariance matrix. The latter hypermodel over values is also represented as a Gaussian distribution with diagonal covariance whose mean is a linear function of an input state abstraction $\mbf{\Phi} \in \bR^{S \times M}$. 
    \item \textbf{PS2-TS} -- Identical parameterization to PS2-IDS except, instead of using variance-IDS, applies Thompson sampling for action selection by drawing a single $Q^\star$ and then acting greedily with respect to the sample. 
    \item \textbf{NoStateAbstraction} -- An implementation of variance-IDS that, instead of learning a state abstraction, directly maintains a single hypermodel of the agent's beliefs over $\mbf{Q}^\star \in \bR^{S \times A}$. 
    \item \textbf{TrueStateAbstraction} -- An implementation of variance-IDS that is given the true state abstraction $\mbf{\Phi}$ a priori and only maintains a hypermodel for learning the corresponding abstract-state values.
    \item \textbf{Independent} -- For multi-task experiments with $T$ distinct tasks, this algorithm maintains $T$ instances of PS2-IDS.
    \item \textbf{Random} -- Selects actions at each timestep uniformly at random.
\end{itemize}

\begin{figure}
\centering
\begin{subfigure}{.5\textwidth}
  \centering
  \includegraphics[width=.9\linewidth]{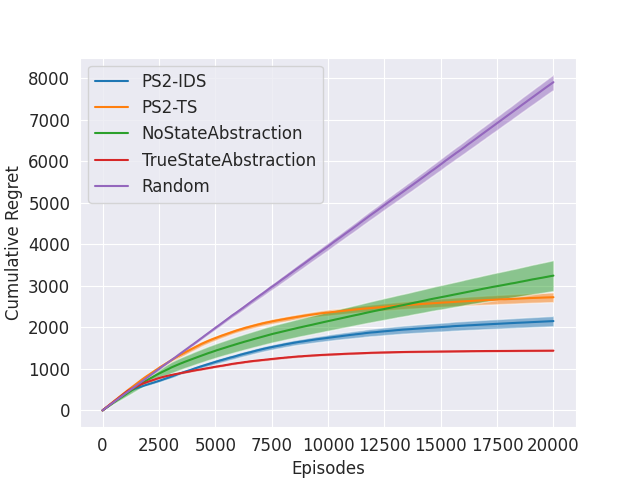}
  \caption{Single-task setting}
  \label{fig:cb_10s10a5r_single}
\end{subfigure}%
\begin{subfigure}{.5\textwidth}
  \centering
  \includegraphics[width=.9\linewidth]{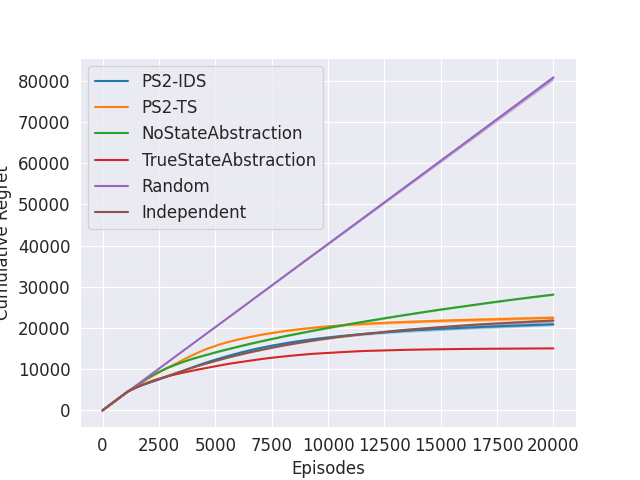}
  \caption{Multi-task setting with 10 tasks.}
  \label{fig:cb_10s10a5r_multi}
\end{subfigure}
\caption{Contextual bandit with $S = 10,A=10,M=5$.}
\label{fig:cb_10s10a5r}
\end{figure}

\begin{figure}
\centering
\begin{subfigure}{.5\textwidth}
  \centering
  \includegraphics[width=.9\linewidth]{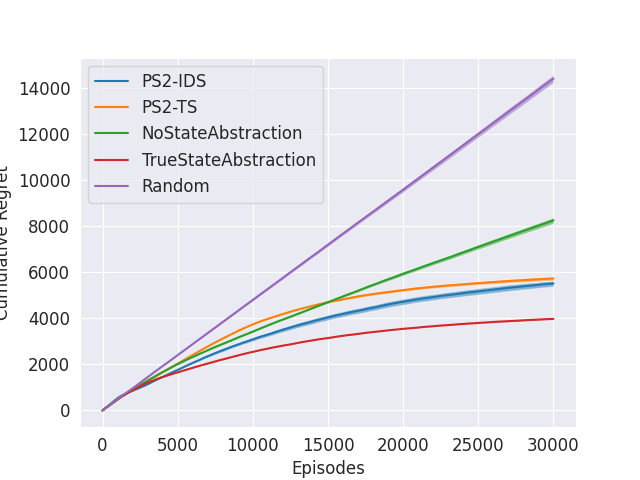}
  \caption{Single-task setting}
  \label{fig:cb_30s30a5r_single}
\end{subfigure}%
\begin{subfigure}{.5\textwidth}
  \centering
  \includegraphics[width=.9\linewidth]{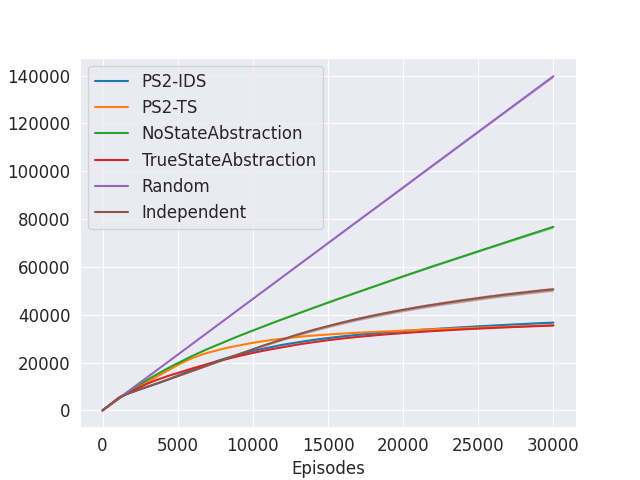}
  \caption{Multi-task setting with 10 tasks.}
  \label{fig:cb_30s30a5r_multi}
\end{subfigure}
\caption{Contextual bandit with $S = 30,A=30,M=5$.}
\label{fig:cb_30s30a5r}
\end{figure}

Figures \ref{fig:cb_10s10a5r} and \ref{fig:cb_30s30a5r} show results for randomly generated contextual bandit problems of two different sizes. Firstly, in the single-task setting, the relationship between the algorithm that has the true state abstraction computed a priori vs. the algorithm that does not pursue any abstraction at all reaffirms existing empirical results that highlight the advantages of leveraging structure in the value function~\citep{abel2016near}. Notably, both PS2 methods are able to achieve performance between these two extremes, identifying the underlying abstraction of the environment to more efficiently arrive at optimal behavior. This point becomes even more apparent in the multi-task setting (Figures \ref{fig:cb_10s10a5r_multi} and \ref{fig:cb_30s30a5r_multi}) where interaction with multiple tasks amplifies the signal provided to the agent for distilling the underlying state abstraction. Due to the small problem size, Figure \ref{fig:cb_10s10a5r_multi} shows little improvement between PS2 and the algorithm that attempts to solve each task in isolation. With a slightly larger problem in Figure \ref{fig:cb_30s30a5r_multi}, however, we observe a substantial improvement in PS2 as it is able to better exploit information from all tasks to capture shared structure. Finally, it is theoretically know that IDS has a stronger performance guarantee than Thompson sampling~\citep{russo2018learning}. Our experiments confirm this relationship with PS2-IDS matching or outperforming Thompson sampling.

\section{Conclusion}
\label{sec:conc}

We have examined state abstractions as a mechanism for facilitating sample-efficient reinforcement learning. While various forms of state abstraction model structure in different components of a MDP, this work places particular focus on those which attempt to leverage structure in the optimal value function. While prior works that study this abstraction type have often been unrealistic in their acquisition of the abstraction itself, we take a Bayesian perspective and leverage insights from past work on provably-efficient, deep exploration. Concretely, we introduce an algorithmic design principle wherein an agent's beliefs over the optimal value function factor into separate posterior distributions over abstract states and abstract-state values, respectively. Empirically, we demonstrate that an agent whose beliefs have been factored in this manner can learn more efficiently than those that attempt to directly estimate values and ignore latent structure altogether. The experiments in this work have been limited to the contextual bandit setting; identifying the right sampling procedure for generating random MDPs with the appropriate structural properties that are conducive for any algorithm, including PS2, to learn state abstractions is an active direction for future work.

\bibliographystyle{plainnat}
\bibliography{references}

\begin{thebibliography}{87}
\providecommand{\natexlab}[1]{#1}
\providecommand{\url}[1]{\texttt{#1}}
\expandafter\ifx\csname urlstyle\endcsname\relax
  \providecommand{\doi}[1]{doi: #1}\else
  \providecommand{\doi}{doi: \begingroup \urlstyle{rm}\Url}\fi

\bibitem[Abel et~al.(2016)Abel, Hershkowitz, and Littman]{abel2016near}
David Abel, D~Ellis Hershkowitz, and Michael~L Littman.
\newblock Near optimal behavior via approximate state abstraction.
\newblock In \emph{Proceedings of the 33rd International Conference on
  International Conference on Machine Learning-Volume 48}, pages 2915--2923,
  2016.

\bibitem[Abel et~al.(2018)Abel, Arumugam, Lehnert, and Littman]{abel2018state}
David Abel, Dilip Arumugam, Lucas Lehnert, and Michael Littman.
\newblock State abstractions for lifelong reinforcement learning.
\newblock In \emph{International Conference on Machine Learning}, pages 10--19,
  2018.

\bibitem[Abel et~al.(2019)Abel, Arumugam, Asadi, Jinnai, Littman, and
  Wong]{abel2019state}
David Abel, Dilip Arumugam, Kavosh Asadi, Yuu Jinnai, Michael~L Littman, and
  Lawson~LS Wong.
\newblock State abstraction as compression in apprenticeship learning.
\newblock In \emph{Proceedings of the AAAI Conference on Artificial
  Intelligence}, volume~33, pages 3134--3142, 2019.

\bibitem[Agarwal et~al.(2020)Agarwal, Kakade, Krishnamurthy, and
  Sun]{agarwal2020flambe}
Alekh Agarwal, Sham Kakade, Akshay Krishnamurthy, and Wen Sun.
\newblock Flambe: Structural complexity and representation learning of low rank
  mdps.
\newblock \emph{arXiv preprint arXiv:2006.10814}, 2020.

\bibitem[Agrawal and Goyal(2012)]{agrawal2012analysis}
Shipra Agrawal and Navin Goyal.
\newblock Analysis of thompson sampling for the multi-armed bandit problem.
\newblock In \emph{Conference on learning theory}, pages 39--1, 2012.

\bibitem[Agrawal and Goyal(2013)]{agrawal2013further}
Shipra Agrawal and Navin Goyal.
\newblock Further optimal regret bounds for thompson sampling.
\newblock In \emph{Artificial intelligence and statistics}, pages 99--107,
  2013.

\bibitem[Agrawal and Jia(2017)]{agrawal2017optimistic}
Shipra Agrawal and Randy Jia.
\newblock Optimistic posterior sampling for reinforcement learning: worst-case
  regret bounds.
\newblock In \emph{Advances in Neural Information Processing Systems}, pages
  1184--1194, 2017.

\bibitem[Auer et~al.(2009)Auer, Jaksch, and Ortner]{auer2009near}
Peter Auer, Thomas Jaksch, and Ronald Ortner.
\newblock Near-optimal regret bounds for reinforcement learning.
\newblock In \emph{Advances in neural information processing systems}, pages
  89--96, 2009.

\bibitem[Barreto et~al.(2016)Barreto, Beirigo, Pineau, and
  Precup]{barreto2016incremental}
Andr{\'e}~MS Barreto, Rafael~L Beirigo, Joelle Pineau, and Doina Precup.
\newblock Incremental stochastic factorization for online reinforcement
  learning.
\newblock In \emph{Thirtieth AAAI Conference on Artificial Intelligence}, 2016.

\bibitem[Baxter(1997)]{baxter1997bayesian}
Jonathan Baxter.
\newblock A bayesian/information theoretic model of learning to learn via
  multiple task sampling.
\newblock \emph{Machine learning}, 28\penalty0 (1):\penalty0 7--39, 1997.

\bibitem[Baxter(2000)]{baxter2000model}
Jonathan Baxter.
\newblock A model of inductive bias learning.
\newblock \emph{Journal of artificial intelligence research}, 12:\penalty0
  149--198, 2000.

\bibitem[Behzadian and Petrik(2018)]{behzadianfeature18}
Bahram Behzadian and Marek Petrik.
\newblock Feature selection by singular value decomposition for reinforcement
  learning.
\newblock In \emph{Proceedings of the ICML Prediction and Generative Modeling
  Workshop}, 2018.

\bibitem[Bellman(1957)]{bellman1957markovian}
Richard Bellman.
\newblock A markovian decision process.
\newblock \emph{Journal of mathematics and mechanics}, pages 679--684, 1957.

\bibitem[Bertsekas et~al.(1988)Bertsekas, Castanon,
  et~al.]{bertsekas1988adaptive}
Dimitri~P Bertsekas, David~A Castanon, et~al.
\newblock Adaptive aggregation methods for infinite horizon dynamic
  programming.
\newblock 1988.

\bibitem[Brafman and Tennenholtz(2002)]{brafman2002r}
Ronen~I Brafman and Moshe Tennenholtz.
\newblock R-max-a general polynomial time algorithm for near-optimal
  reinforcement learning.
\newblock \emph{Journal of Machine Learning Research}, 3\penalty0
  (Oct):\penalty0 213--231, 2002.

\bibitem[Bubeck and Cesa-Bianchi(2012)]{bubeck2012regret}
S{\'e}bastien Bubeck and Nicolo Cesa-Bianchi.
\newblock Regret analysis of stochastic and nonstochastic multi-armed bandit
  problems.
\newblock \emph{arXiv preprint arXiv:1204.5721}, 2012.

\bibitem[Calandriello et~al.(2014)Calandriello, Lazaric, and
  Restelli]{calandriello2014sparse}
Daniele Calandriello, Alessandro Lazaric, and Marcello Restelli.
\newblock Sparse multi-task reinforcement learning.
\newblock In \emph{Advances in Neural Information Processing Systems}, pages
  819--827, 2014.

\bibitem[Cand{\`e}s and Plan(2010)]{candes2010matrix}
Emmanuel~J Cand{\`e}s and Yaniv Plan.
\newblock Matrix completion with noise.
\newblock \emph{Proceedings of the IEEE}, 98\penalty0 (6):\penalty0 925--936,
  2010.

\bibitem[Cand{\`e}s and Recht(2009)]{candes2009exact}
Emmanuel~J Cand{\`e}s and Benjamin Recht.
\newblock Exact matrix completion via convex optimization.
\newblock \emph{Foundations of Computational mathematics}, 9\penalty0
  (6):\penalty0 717, 2009.

\bibitem[Cand{\`e}s and Tao(2010)]{candes2010power}
Emmanuel~J Cand{\`e}s and Terence Tao.
\newblock The power of convex relaxation: Near-optimal matrix completion.
\newblock \emph{IEEE Transactions on Information Theory}, 56\penalty0
  (5):\penalty0 2053--2080, 2010.

\bibitem[Caruana(1997)]{caruana1997multitask}
Rich Caruana.
\newblock Multitask learning.
\newblock \emph{Machine learning}, 28\penalty0 (1):\penalty0 41--75, 1997.

\bibitem[Chapelle and Li(2011)]{chapelle2011empirical}
Olivier Chapelle and Lihong Li.
\newblock An empirical evaluation of thompson sampling.
\newblock In \emph{Advances in neural information processing systems}, pages
  2249--2257, 2011.

\bibitem[Chen and Wainwright(2015)]{chen2015fast}
Yudong Chen and Martin~J Wainwright.
\newblock Fast low-rank estimation by projected gradient descent: General
  statistical and algorithmic guarantees.
\newblock \emph{arXiv preprint arXiv:1509.03025}, 2015.

\bibitem[Chistov and Grigor'Ev(1984)]{chistov1984complexity}
Alexander~L Chistov and D~Yu Grigor'Ev.
\newblock Complexity of quantifier elimination in the theory of algebraically
  closed fields.
\newblock In \emph{International Symposium on Mathematical Foundations of
  Computer Science}, pages 17--31. Springer, 1984.

\bibitem[Dean and Givan(1997)]{dean1997model}
Thomas Dean and Robert Givan.
\newblock Model minimization in markov decision processes.
\newblock In \emph{Proceedings of the AAAI Conference on Artificial
  Intelligence}, pages 106--111. AAAI Press, 1997.

\bibitem[D'Eramo et~al.(2019)D'Eramo, Tateo, Bonarini, Restelli, and
  Peters]{d2019sharing}
Carlo D'Eramo, Davide Tateo, Andrea Bonarini, Marcello Restelli, and Jan
  Peters.
\newblock Sharing knowledge in multi-task deep reinforcement learning.
\newblock In \emph{International Conference on Learning Representations}, 2019.

\bibitem[Dong et~al.(2019)Dong, Van~Roy, and Zhou]{dong2019provably}
Shi Dong, Benjamin Van~Roy, and Zhengyuan Zhou.
\newblock Provably efficient reinforcement learning with aggregated states.
\newblock \emph{arXiv preprint arXiv:1912.06366}, 2019.

\bibitem[Donoho and Stodden(2004)]{donoho2004does}
David Donoho and Victoria Stodden.
\newblock When does non-negative matrix factorization give a correct
  decomposition into parts?
\newblock In \emph{Advances in neural information processing systems}, pages
  1141--1148, 2004.

\bibitem[Du et~al.(2019)Du, Krishnamurthy, Jiang, Agarwal, Dudik, and
  Langford]{du2019provably}
Simon Du, Akshay Krishnamurthy, Nan Jiang, Alekh Agarwal, Miroslav Dudik, and
  John Langford.
\newblock Provably efficient rl with rich observations via latent state
  decoding.
\newblock In \emph{International Conference on Machine Learning}, pages
  1665--1674, 2019.

\bibitem[Duan et~al.(2019)Duan, Ke, and Wang]{duan2019state}
Yaqi Duan, Tracy Ke, and Mengdi Wang.
\newblock State aggregation learning from markov transition data.
\newblock In \emph{Advances in Neural Information Processing Systems}, pages
  4488--4497, 2019.

\bibitem[Dwaracherla et~al.(2020)Dwaracherla, Lu, Ibrahimi, Osband, Wen, and
  Van~Roy]{dwaracherla2020hypermodels}
Vikranth Dwaracherla, Xiuyuan Lu, Morteza Ibrahimi, Ian Osband, Zheng Wen, and
  Benjamin Van~Roy.
\newblock Hypermodels for exploration.
\newblock In \emph{International Conference on Learning Representations}, 2020.

\bibitem[Ferns et~al.(2004)Ferns, Panangaden, and Precup]{ferns2004metrics}
Norm Ferns, Prakash Panangaden, and Doina Precup.
\newblock Metrics for finite markov decision processes.
\newblock 2004.

\bibitem[Ferns et~al.(2012)Ferns, Castro, Precup, and
  Panangaden]{ferns2012methods}
Norman Ferns, Pablo~Samuel Castro, Doina Precup, and Prakash Panangaden.
\newblock Methods for computing state similarity in markov decision processes.
\newblock \emph{arXiv preprint arXiv:1206.6836}, 2012.

\bibitem[Gunasekar et~al.(2013)Gunasekar, Acharya, Gaur, and
  Ghosh]{gunasekar2013noisy}
Suriya Gunasekar, Ayan Acharya, Neeraj Gaur, and Joydeep Ghosh.
\newblock Noisy matrix completion using alternating minimization.
\newblock In \emph{Joint European Conference on Machine Learning and Knowledge
  Discovery in Databases}, pages 194--209. Springer, 2013.

\bibitem[Hardt(2014)]{hardt2014understanding}
Moritz Hardt.
\newblock Understanding alternating minimization for matrix completion.
\newblock In \emph{2014 IEEE 55th Annual Symposium on Foundations of Computer
  Science}, pages 651--660. IEEE, 2014.

\bibitem[Hardt et~al.(2014)Hardt, Meka, Raghavendra, and
  Weitz]{hardt2014computational}
Moritz Hardt, Raghu Meka, Prasad Raghavendra, and Benjamin Weitz.
\newblock Computational limits for matrix completion.
\newblock In \emph{Conference on Learning Theory}, pages 703--725, 2014.

\bibitem[Hazan et~al.(2019)Hazan, Kakade, Singh, and
  Van~Soest]{hazan2019provably}
Elad Hazan, Sham Kakade, Karan Singh, and Abby Van~Soest.
\newblock Provably efficient maximum entropy exploration.
\newblock In \emph{International Conference on Machine Learning}, pages
  2681--2691, 2019.

\bibitem[Jaderberg et~al.(2016)Jaderberg, Mnih, Czarnecki, Schaul, Leibo,
  Silver, and Kavukcuoglu]{jaderberg2016reinforcement}
Max Jaderberg, Volodymyr Mnih, Wojciech~Marian Czarnecki, Tom Schaul, Joel~Z
  Leibo, David Silver, and Koray Kavukcuoglu.
\newblock Reinforcement learning with unsupervised auxiliary tasks.
\newblock \emph{arXiv preprint arXiv:1611.05397}, 2016.

\bibitem[Jain et~al.(2013)Jain, Netrapalli, and Sanghavi]{jain2013low}
Prateek Jain, Praneeth Netrapalli, and Sujay Sanghavi.
\newblock Low-rank matrix completion using alternating minimization.
\newblock In \emph{Proceedings of the forty-fifth annual ACM symposium on
  Theory of computing}, pages 665--674, 2013.

\bibitem[Jiang et~al.(2015)Jiang, Kulesza, and Singh]{jiang2015abstraction}
Nan Jiang, Alex Kulesza, and Satinder Singh.
\newblock Abstraction selection in model-based reinforcement learning.
\newblock In \emph{International Conference on Machine Learning}, pages
  179--188, 2015.

\bibitem[Jin et~al.(2020)Jin, Krishnamurthy, Simchowitz, and Yu]{jin2020reward}
Chi Jin, Akshay Krishnamurthy, Max Simchowitz, and Tiancheng Yu.
\newblock Reward-free exploration for reinforcement learning.
\newblock \emph{arXiv preprint arXiv:2002.02794}, 2020.

\bibitem[Jong and Stone(2005)]{jong2005state}
Nicholas~K Jong and Peter Stone.
\newblock State abstraction discovery from irrelevant state variables.
\newblock In \emph{Proceedings of the 19th international joint conference on
  Artificial intelligence}, pages 752--757, 2005.

\bibitem[Kakade(2003)]{kakade2003sample}
Sham~Machandranath Kakade.
\newblock \emph{On the sample complexity of reinforcement learning}.
\newblock PhD thesis, 2003.

\bibitem[Kearns and Singh(2002)]{kearns2002near}
Michael Kearns and Satinder Singh.
\newblock Near-optimal reinforcement learning in polynomial time.
\newblock \emph{Machine learning}, 49\penalty0 (2-3):\penalty0 209--232, 2002.

\bibitem[Keshavan et~al.(2010{\natexlab{a}})Keshavan, Montanari, and
  Oh]{keshavan2010matrix}
Raghunandan~H Keshavan, Andrea Montanari, and Sewoong Oh.
\newblock Matrix completion from a few entries.
\newblock \emph{IEEE transactions on information theory}, 56\penalty0
  (6):\penalty0 2980--2998, 2010{\natexlab{a}}.

\bibitem[Keshavan et~al.(2010{\natexlab{b}})Keshavan, Montanari, and
  Oh]{keshavan2010matrixnoisy}
Raghunandan~H Keshavan, Andrea Montanari, and Sewoong Oh.
\newblock Matrix completion from noisy entries.
\newblock \emph{Journal of Machine Learning Research}, 11\penalty0
  (Jul):\penalty0 2057--2078, 2010{\natexlab{b}}.

\bibitem[Kingma and Ba(2014)]{kingma2014adam}
Diederik~P Kingma and Jimmy Ba.
\newblock Adam: A method for stochastic optimization.
\newblock \emph{arXiv preprint arXiv:1412.6980}, 2014.

\bibitem[Konidaris(2019)]{konidaris2019necessity}
George Konidaris.
\newblock On the necessity of abstraction.
\newblock \emph{Current opinion in behavioral sciences}, 29:\penalty0 1--7,
  2019.

\bibitem[Koren et~al.(2009)Koren, Bell, and Volinsky]{koren2009matrix}
Yehuda Koren, Robert Bell, and Chris Volinsky.
\newblock Matrix factorization techniques for recommender systems.
\newblock \emph{Computer}, 42\penalty0 (8):\penalty0 30--37, 2009.

\bibitem[Krishnamurthy and Singh(2013)]{krishnamurthy2013low}
Akshay Krishnamurthy and Aarti Singh.
\newblock Low-rank matrix and tensor completion via adaptive sampling.
\newblock In \emph{Advances in neural information processing systems}, pages
  836--844, 2013.

\bibitem[Krishnamurthy and Singh(2014)]{krishnamurthy2014power}
Akshay Krishnamurthy and Aarti Singh.
\newblock On the power of adaptivity in matrix completion and approximation.
\newblock \emph{arXiv preprint arXiv:1407.3619}, 2014.

\bibitem[Lake et~al.(2017)Lake, Ullman, Tenenbaum, and
  Gershman]{lake2017building}
Brenden~M Lake, Tomer~D Ullman, Joshua~B Tenenbaum, and Samuel~J Gershman.
\newblock Building machines that learn and think like people.
\newblock \emph{Behavioral and brain sciences}, 40, 2017.

\bibitem[Li et~al.(2006)Li, Walsh, and Littman]{li2006towards}
Lihong Li, Thomas~J. Walsh, and Michael~L. Littman.
\newblock Towards a unified theory of state abstraction for {MDP}s.
\newblock In \emph{Proceedings of the International Symposium on Artificial
  Intelligence and Mathematics}, 2006.

\bibitem[Lin(1992)]{lin1992self}
Long-Ji Lin.
\newblock Self-improving reactive agents based on reinforcement learning,
  planning and teaching.
\newblock \emph{Machine learning}, 8\penalty0 (3-4):\penalty0 293--321, 1992.

\bibitem[Lu and Van~Roy(2017)]{lu2017ensemble}
Xiuyuan Lu and Benjamin Van~Roy.
\newblock Ensemble sampling.
\newblock In \emph{Advances in neural information processing systems}, pages
  3258--3266, 2017.

\bibitem[Misra et~al.(2019)Misra, Henaff, Krishnamurthy, and
  Langford]{misra2019kinematic}
Dipendra Misra, Mikael Henaff, Akshay Krishnamurthy, and John Langford.
\newblock Kinematic state abstraction and provably efficient rich-observation
  reinforcement learning.
\newblock \emph{arXiv preprint arXiv:1911.05815}, 2019.

\bibitem[Mnih et~al.(2015)Mnih, Kavukcuoglu, Silver, Rusu, Veness, Bellemare,
  Graves, Riedmiller, Fidjeland, Ostrovski, et~al.]{mnih2015human}
Volodymyr Mnih, Koray Kavukcuoglu, David Silver, Andrei~A Rusu, Joel Veness,
  Marc~G Bellemare, Alex Graves, Martin Riedmiller, Andreas~K Fidjeland, Georg
  Ostrovski, et~al.
\newblock Human-level control through deep reinforcement learning.
\newblock \emph{nature}, 518\penalty0 (7540):\penalty0 529--533, 2015.

\bibitem[M{\"u}ller(1997)]{muller1997integral}
Alfred M{\"u}ller.
\newblock Integral probability metrics and their generating classes of
  functions.
\newblock \emph{Advances in Applied Probability}, 29\penalty0 (2):\penalty0
  429--443, 1997.

\bibitem[Osband and Van~Roy(2017)]{osband2017posterior}
Ian Osband and Benjamin Van~Roy.
\newblock Why is posterior sampling better than optimism for reinforcement
  learning?
\newblock In \emph{International Conference on Machine Learning}, pages
  2701--2710, 2017.

\bibitem[Osband et~al.(2016{\natexlab{a}})Osband, Blundell, Pritzel, and
  Van~Roy]{osband2016deep}
Ian Osband, Charles Blundell, Alexander Pritzel, and Benjamin Van~Roy.
\newblock Deep exploration via bootstrapped dqn.
\newblock In \emph{Advances in neural information processing systems}, pages
  4026--4034, 2016{\natexlab{a}}.

\bibitem[Osband et~al.(2016{\natexlab{b}})Osband, Van~Roy, and
  Wen]{osband2016generalization}
Ian Osband, Benjamin Van~Roy, and Zheng Wen.
\newblock Generalization and exploration via randomized value functions.
\newblock In \emph{International Conference on Machine Learning}, pages
  2377--2386, 2016{\natexlab{b}}.

\bibitem[Osband et~al.(2019)Osband, Van~Roy, Russo, and Wen]{osband2019deep}
Ian Osband, Benjamin Van~Roy, Daniel~J Russo, and Zheng Wen.
\newblock Deep exploration via randomized value functions.
\newblock \emph{Journal of Machine Learning Research}, 20\penalty0
  (124):\penalty0 1--62, 2019.

\bibitem[O’Donoghue et~al.(2018)O’Donoghue, Osband, Munos, and
  Mnih]{o2018uncertainty}
Brendan O’Donoghue, Ian Osband, Remi Munos, and Volodymyr Mnih.
\newblock The uncertainty bellman equation and exploration.
\newblock In \emph{International Conference on Machine Learning}, pages
  3836--3845, 2018.

\bibitem[Pinsker(1960)]{Pinsker1960InformationAI}
Mark~Semenovich Pinsker.
\newblock Information and information stability of random variables and
  processes.
\newblock 1960.

\bibitem[Puterman(1994)]{Puterman94}
Martin~L. Puterman.
\newblock \emph{Markov Decision Processes---Discrete Stochastic Dynamic
  Programming}.
\newblock John Wiley \& Sons, Inc., New York, NY, 1994.

\bibitem[Recht(2011)]{recht2011simpler}
Benjamin Recht.
\newblock A simpler approach to matrix completion.
\newblock \emph{Journal of Machine Learning Research}, 12\penalty0
  (Dec):\penalty0 3413--3430, 2011.

\bibitem[Recht and R{\'e}(2013)]{recht2013parallel}
Benjamin Recht and Christopher R{\'e}.
\newblock Parallel stochastic gradient algorithms for large-scale matrix
  completion.
\newblock \emph{Mathematical Programming Computation}, 5\penalty0 (2):\penalty0
  201--226, 2013.

\bibitem[Russo(2020)]{russo2020approximation}
Daniel Russo.
\newblock Approximation benefits of policy gradient methods with aggregated
  states.
\newblock \emph{arXiv preprint arXiv:2007.11684}, 2020.

\bibitem[Russo and Van~Roy(2016)]{russo2016information}
Daniel Russo and Benjamin Van~Roy.
\newblock An information-theoretic analysis of thompson sampling.
\newblock \emph{The Journal of Machine Learning Research}, 17\penalty0
  (1):\penalty0 2442--2471, 2016.

\bibitem[Russo and Van~Roy(2018)]{russo2018learning}
Daniel Russo and Benjamin Van~Roy.
\newblock Learning to optimize via information-directed sampling.
\newblock \emph{Operations Research}, 66\penalty0 (1):\penalty0 230--252, 2018.

\bibitem[Shah et~al.(2020)Shah, Song, Xu, and Yang]{shah2020sample}
Devavrat Shah, Dogyoon Song, Zhi Xu, and Yuzhe Yang.
\newblock Sample efficient reinforcement learning via low-rank matrix
  estimation.
\newblock \emph{arXiv preprint arXiv:2006.06135}, 2020.

\bibitem[Shannon(1959)]{shannon1959coding}
Claude~E. Shannon.
\newblock Coding theorems for a discrete source with a fidelity criterion.
\newblock \emph{IRE Nat. Conv. Rec., March 1959}, 4:\penalty0 142--163, 1959.

\bibitem[Singh et~al.(1995)Singh, Jaakkola, and Jordan]{singh1995reinforcement}
Satinder~P Singh, Tommi Jaakkola, and Michael~I Jordan.
\newblock Reinforcement learning with soft state aggregation.
\newblock In \emph{Advances in neural information processing systems}, pages
  361--368, 1995.

\bibitem[Strehl et~al.(2009)Strehl, Li, and Littman]{strehl2009reinforcement}
Alexander~L Strehl, Lihong Li, and Michael~L Littman.
\newblock Reinforcement learning in finite mdps: Pac analysis.
\newblock \emph{Journal of Machine Learning Research}, 10\penalty0
  (Nov):\penalty0 2413--2444, 2009.

\bibitem[Sutton(1988)]{sutton1988learning}
Richard~S Sutton.
\newblock Learning to predict by the methods of temporal differences.
\newblock \emph{Machine learning}, 3\penalty0 (1):\penalty0 9--44, 1988.

\bibitem[Sutton and Barto(1998)]{sutton1998introduction}
Richard~S Sutton and Andrew~G Barto.
\newblock Introduction to reinforcement learning.
\newblock 1998.

\bibitem[Sutton et~al.(2000)Sutton, McAllester, Singh, and
  Mansour]{sutton2000policy}
Richard~S Sutton, David~A McAllester, Satinder~P Singh, and Yishay Mansour.
\newblock Policy gradient methods for reinforcement learning with function
  approximation.
\newblock In \emph{Advances in neural information processing systems}, pages
  1057--1063, 2000.

\bibitem[Sutton et~al.(2011)Sutton, Modayil, Delp, Degris, Pilarski, White, and
  Precup]{sutton2011horde}
Richard~S Sutton, Joseph Modayil, Michael Delp, Thomas Degris, Patrick~M
  Pilarski, Adam White, and Doina Precup.
\newblock Horde: A scalable real-time architecture for learning knowledge from
  unsupervised sensorimotor interaction.
\newblock In \emph{The 10th International Conference on Autonomous Agents and
  Multiagent Systems-Volume 2}, pages 761--768, 2011.

\bibitem[Thompson(1933)]{thompson1933likelihood}
William~R Thompson.
\newblock On the likelihood that one unknown probability exceeds another in
  view of the evidence of two samples.
\newblock \emph{Biometrika}, 25\penalty0 (3/4):\penalty0 285--294, 1933.

\bibitem[Tsitsiklis and Van~Roy(1996)]{tsitsiklis1996feature}
John~N Tsitsiklis and Benjamin Van~Roy.
\newblock Feature-based methods for large scale dynamic programming.
\newblock \emph{Machine Learning}, 22\penalty0 (1-3):\penalty0 59--94, 1996.

\bibitem[Van~Roy(2006)]{van2006performance}
Benjamin Van~Roy.
\newblock Performance loss bounds for approximate value iteration with state
  aggregation.
\newblock \emph{Mathematics of Operations Research}, 31\penalty0 (2):\penalty0
  234--244, 2006.

\bibitem[Watkins and Dayan(1992)]{watkins1992q}
Christopher~JCH Watkins and Peter Dayan.
\newblock Q-learning.
\newblock \emph{Machine learning}, 8\penalty0 (3-4):\penalty0 279--292, 1992.

\bibitem[Whitt(1978)]{whitt1978approximations}
Ward Whitt.
\newblock Approximations of dynamic programs, i.
\newblock \emph{Mathematics of Operations Research}, 3\penalty0 (3):\penalty0
  231--243, 1978.

\bibitem[Williams(1992)]{williams1992simple}
Ronald~J Williams.
\newblock Simple statistical gradient-following algorithms for connectionist
  reinforcement learning.
\newblock \emph{Machine learning}, 8\penalty0 (3-4):\penalty0 229--256, 1992.

\bibitem[Yang and Wang(2019{\natexlab{a}})]{yang2019sample}
Lin Yang and Mengdi Wang.
\newblock Sample-optimal parametric q-learning using linearly additive
  features.
\newblock In \emph{International Conference on Machine Learning}, pages
  6995--7004, 2019{\natexlab{a}}.

\bibitem[Yang and Wang(2019{\natexlab{b}})]{yang2019reinforcement}
Lin~F Yang and Mengdi Wang.
\newblock Reinforcement learning in feature space: Matrix bandit, kernels, and
  regret bound.
\newblock \emph{arXiv preprint arXiv:1905.10389}, 2019{\natexlab{b}}.

\bibitem[Zhang and Wang(2019)]{zhang2019spectral}
Anru Zhang and Mengdi Wang.
\newblock Spectral state compression of markov processes.
\newblock \emph{IEEE Transactions on Information Theory}, 66\penalty0
  (5):\penalty0 3202--3231, 2019.

\end{thebibliography}

\appendix

\section{Related Work}
\label{sec:related}

As previously discussed, this paper falls in with a long, rich line of work on state abstraction in reinforcement learning~\citep{whitt1978approximations,bertsekas1988adaptive,dean1997model,ferns2004metrics,jong2005state,li2006towards,van2006performance,ferns2012methods,jiang2015abstraction,abel2016near,abel2019state,dong2019provably,du2019provably,misra2019kinematic}. Notably, this work is concerned with how an agent may incrementally learn a state abstraction that capitalizes on latent structure in the optimal value function~\citep{bertsekas1988adaptive,van2006performance,li2006towards,abel2016near,abel2018state,abel2019state,dong2019provably}.  \citet{bertsekas1988adaptive} focus on adaptively synthesizing state aggregations based on Bellman-error residuals. \citet{van2006performance} examines and provides performance guarantees on approximate value iteration under the provision of a particular state aggregation, leaving open the question of how to dynamically abstract states and maintain performance guarantees. \citet{li2006towards} offer a unified perspective on a broad array of state abstraction types, showcasing how state abstraction based on $Q^\star$-similarity preserves the optimal policy. Later, \citet{abel2016near} generalize this to the case of approximate state abstraction, highlighting the approximation parameter as a knob for weighing state-space compression against value loss. \citet{abel2018state} examine the lifelong learning setting where $Q^\star$ is computed exactly for some number of MDPs before a $Q^\star$-similarity state abstraction is then applied for the remainder of the task distribution. \citet{abel2019state} formalize this intuition using tools from rate-distortion theory~\citep{shannon1959coding}, but restrict focus to the apprenticeship learning setting. Common to all of these works is the lack of a practical, scalable algorithm for jointly learning the state abstraction and corresponding abstract-state values, without knowledge of $Q^\star$; our work rectifies this and offers one such approach.

Our work is also intimately related to the problem of low-rank matrix completion or factorization. While the general problem of low-rank matrix completion is underspecified and known to be NP-hard~\citep{chistov1984complexity,hardt2014computational}, a large body of prior work identifies sufficient conditions for designing provably-efficient factorization algorithms~\citep{candes2009exact,candes2010power,candes2010matrix,keshavan2010matrix,keshavan2010matrixnoisy,recht2011simpler}. The algorithm presented in this work aligns with gradient-descent based approaches for iteratively optimizing the latent factors $\mbf{\Phi},\mbf{\Psi}$ which, despite their weaker sample complexity guarantees~\citep{jain2013low,gunasekar2013noisy,hardt2014understanding,chen2015fast}, are simple, scalable, and widely deployed in practice~\citep{recht2013parallel,koren2009matrix}. Unlike the standard formulation of the low-rank matrix completion problem, our focus on the sequential decision-making setting more closely aligns with adaptive-sampling approaches to matrix completion~\citep{krishnamurthy2013low,krishnamurthy2014power} which are known to enjoy better sample complexity guarantees.

Several papers adopt a matrix factorization perspective for state abstraction in reinforcement learning~\citep{barreto2016incremental,behzadianfeature18,duan2019state,zhang2019spectral,yang2019reinforcement,yang2019sample,agarwal2020flambe}; crucially, however, these works opt for computing a factorization of the transition function, rather than the value function. Again, we suspect that this preference stems from the immediate inaccessibility of $Q^\star$, a fact that we show need not be an obstacle when adopting a Bayesian view of efficient exploration. Moreover, leveraging such a Bayesian approach to exploration potentially avoids the known pitfalls~\citep{osband2017posterior} of alternative methods that employ optimism in the face of uncertainty~\citep{yang2019reinforcement}. Most related to this work is the approach of \citet{shah2020sample} who do in fact aim to learn the singular value decomposition of $Q^\star$ and take advantage of low-rank structure; important differences from this work include a focus on MDPs with continuous state-action spaces along with assumptions on the Lipschitz continuity of $Q^\star$, access to a generative model for sampling transitions, and access to anchor states (states that are representative of each latent abstract state~\citep{donoho2004does}) for heuristically guiding exploration/data collection. In contrast, this work is concerned with discrete-action MDPs (though extensions to the continuous-control setting are a natural future direction) and makes no assumptions on $Q^\star$ beyond being low rank (Assumption \ref{asm:low_rank}). 

For understanding when it is prudent for an agent to pursue latent task structure in the form of a state abstraction, we find the multi-task setting to be a natural candidate. In the context of the low-rank matrix factorization outlined in the previous section, this amounts to asserting that the optimal value functions of all tasks share a common, latent factor $\mbf{\Phi}$ while each individual task $t$ also yields a specific matrix of abstract-state values, $\mbf{\Psi}_t$. Consequently, an agent interacting to solve all tasks in parallel can greatly benefit from synthesizing shared task structure. Various prior works already assess such benefits of multiple tasks for supervised learning~\citep{caruana1997multitask,baxter1997bayesian,baxter2000model}. While work that formally develops this connection for reinforcement learning is still nascent~\citep{d2019sharing,calandriello2014sparse}, empirical examples of this phenomenon are well-established~\citep{sutton2011horde,jaderberg2016reinforcement}. Our work can be seen as a simple mathematical model for studying this phenomenon in the Bayesian RL setting.

\section{Background}
\label{sec:back}

In this section, we provide background on provably-efficient approaches to addressing the exploration-exploitation trade-off in sequential decision-making problems. The mechanisms employed by these approaches for representing uncertainty will play a central role in our algorithm for learning a value-based state abstraction.


\subsection{Information-Directed Sampling}
\label{sec:ids}

A central challenge that all sequential decision-making agents must confront is that of exploration; an agent must strike a delicate balance between acquiring new knowledge in the hope of improving future performance or capitalizing on the information it has acquired thus far. Early results for provably-efficient reinforcement-learning algorithms (designated PAC-MDP~\citep{strehl2009reinforcement}) hinge on the sufficiency of an agent's exploration strategy for fully exploring the MDP~\citep{kakade2003sample}, typically based on a principle of optimism in the face of uncertainty~\citep{kearns2002near,brafman2002r,bubeck2012regret}. In recent years, exploration techniques that facilitate stronger theoretical guarantees have come about by leveraging estimates of an agent's epistemic uncertainty or uncertainty stemming from parameter estimation (rather than the aleatoric uncertainty driven by stochasticity in data)~\citep{chapelle2011empirical,russo2016information,osband2016deep,agrawal2017optimistic,o2018uncertainty,osband2019deep}. 

In the context of multi-armed bandit problems, an agent maintains uncertainty over the individual reward or payoff functions at each arm. With uncertainty in the rewards of all arms driving uncertainty over optimal actions, one choice is for the agent to employ an exploration scheme based on Thompson sampling (TS)~\citep{thompson1933likelihood,agrawal2012analysis,agrawal2013further,russo2016information} whereby an agent acts optimally with respect to a single sample drawn from its posterior beliefs at each time period. This idea naturally scales to the full reinforcement-learning scenario wherein posterior beliefs are maintained over the optimal action-value function $Q^\star$~\citep{osband2016deep,o2018uncertainty,osband2019deep}.

A significant advance on the aforementioned exploration scheme is the algorithmic design principle known as information-directed sampling (IDS)~\citep{russo2018learning}. While previous approaches follow suit with Thompson sampling and act optimally according to posterior samples, IDS algorithms execute a policy at each time period that solves the following minimization problem
\begin{align*}
    \pi_h &= \min\limits_{\pi \in \Delta(\mc{A})} \frac{(\bE[\Delta_h(\pi)])^2
    }{\mc{I}(\theta;(S_h,A_h) | \mathcal{E}_{h-1})}
\end{align*}

where $\bE[\Delta_h(\pi)]$ denotes the expected regret of policy $\pi$ under the agent's current posterior beliefs and $\mc{I}(\theta;(S_h,A_h) | \mathcal{E}_{h-1})$ denotes the expected information gain between the behavior at timestep $h$ and the environment parameters $\theta$, conditioned on the history of episodes collected thus far, $\mc{E}_{h-1}$. For a multi-armed bandit problem, $\theta$ reflects the reward or payoffs at each arm whereas, for a MDP, $\theta$ captures the environment transition function and reward function. IDS embodies an intuitive principle that, rather than being biased exclusively towards the optimal action of one posterior sample, an agent should be incentivized to take one or more suboptimal actions so long as they are informative and revelatory of the underlying environment, $\theta$. Here, the information gain term in the denominator of the information ratio above quantifies this level of informativity, weighing it against the agent's desire to minimize regret over its lifetime. Again, while simple and powerful, IDS is only a design principle to guide the development of practical, efficient algorithms. In the next section, we discuss a deliberate choice of how to represent an agent's posterior beliefs that yields a concrete instantiation of IDS for our algorithm.




\subsection{Variance-IDS}
Previously, we specify how hypermodels can be used to maintain approximate posterior distributions over state abstraction and abstract state values. The final outstanding component that must be specified is how these approximate posterior distributions can be folded into an algorithm that instantiates IDS as the core exploration strategy. To do this, we leverage variance-IDS as introduced in \citet{russo2018learning} and as specified for hypermodels in \citet{dwaracherla2020hypermodels}. For clarity, we present the derivation of variance-IDS.

Recall the definition of the mutual information between two random variables $X,Y$:
\begin{align*}
    \mc{I}(X;Y) &= \bE_X[\kl{p(Y|X)}{p(Y)}]
\end{align*}

Noting the definition contains a Kullback-Leibler (KL) divergence term, we also define Pinkser's inequality~\citep{Pinsker1960InformationAI}
\begin{align*}
    D_\text{TV}(p(X)||q(X)) &\leq \sqrt{\frac{1}{2}\kl{p(X)}{q(X)}}
\end{align*}
where $D_\text{TV}(p(X)||q(X))$ denotes the total variation distance between distributions $p,q$. Lastly, we note that the total variation distance is an integral probability metric (IPM)~\citep{muller1997integral} which, for a random variable $X$ with support $\mc{X}$, is defined as:
\begin{align*}
    D_\text{TV}(p(X)||q(X)) &= \sup\limits_{\substack{f:\mc{X} \mapsto \bR \\ ||f||_\infty \leq 1}} \bE_{p(X)}[f(X)] - \bE_{q(X)}[f(X)]
\end{align*}
where the supremum is taken with respect to all witness functions $f:\mc{X} \mapsto \bR$ with infinity norm bounded by 1. Putting all the pieces together, we recall that IDS balances regret minimization with the selection of informative actions, where informativeness of agent behavior is measured by the (conditional) mutual information\footnote{Although not denoted here for clarity, this is mutual information term is conditioned on the entire history of all past interactions up to this point.} between the state-action pair observed at timestep $h$ and the true environment parameters $\theta$,  $\mc{I}(\theta;S_h,A_h)$. Following \citet{russo2018learning}, this term can be lower bounded as follows:

\begin{align*}
    \mc{I}(\theta;S_h,A_h) &= \bE_{\theta}[D_{KL}(p(S_h,A_h|\theta)||p(S_h,A_h))] \\
    &\geq 2 \bE_{\theta}[(D_{TV}(p(S_h,A_h|\theta)||p(S_h,A_h)))^2] \\
    &= 2 \bE_{\theta}[(\sup\limits_{\substack{f:\mc{S} \times \mc{A} \mapsto \bR \\ ||f||_\infty \leq 1}}( \bE_{p(S_h,A_h|\theta)}[f(S_h,A_h)] - \bE_{p(S_h,A_h)}[f(S_h,A_h)])^2] \\
    &\geq 2 \bE_{\theta}[(\bE[Q^\star(S_h,A_h) | \theta] - \bE[Q^\star_h(S_h,A_h)])^2] \\
    &= 2 \bV[\bE[Q^\star(S_h,A_h)|\theta]]
\end{align*}

where the steps follow from the definition of mutual information, Pinsker's inequality, the IPM form of the total variation distance, the definition of supremum, and the definition of variance. Crucially, and just as with the original mutual information term, all of the expectations and variances above are conditioned on the past history of interactions up to this timestep. Thus, the above shows that the information gain at timestep $h$ is lower bounded by the extent to which $Q^\star$-values produced by action $a$ vary under the identity of the environment parameters $\theta$. Consequently, those actions with high variance in $Q^\star$-values under the agent's current posterior beliefs are deemed to be the most informative.

\section{Posterior State-Abstraction Sampling Algorithm}

\begin{algorithm}[H]
\SetAlgoLined
\KwData{Reference distributions $p_{z_\phi}, p_{z_\psi} = \mc{N}(0,I)$, Minibatch size $m$, Learning rate $\alpha$, Regularization parameters $\lambda$, Noise variances $\sigma^2_\phi$, $\sigma^2_\psi$}

Initialize $\mf{H}^\phi_\nu, \mf{H}^\psi_\nu$ 

$\mc{D} \leftarrow \emptyset$

\For{t $1,\ldots,T$}{
    \For{h $1,\ldots,H$}{
        Sample $Z = \{(z_\phi, z_\psi)_1, \ldots, (z_\phi, z_\psi)_K\}, z_\phi \sim p_{z_\phi}(\cdot), z_\psi \sim p_{z_\psi}(\cdot)$
        
        $\hat{Q}^\star_{(z_\phi,z_\psi)}(s,a) \triangleq \langle \mf{H}^\phi_\nu(z_\phi)(s),\mf{H}^\psi_\nu(z_\psi,\mf{H}^\phi_\nu(z_\phi))(a)\rangle$
        
        $\hat{\Delta}_a(s) \triangleq \frac{1}{|Z|} \sum\limits_{i=1}^K \max\limits_{a^\star \in \mc{A}} \hat{Q}^\star_{(z_\phi,z_\psi)_i}(s,a^\star) - \hat{Q}^\star_{(z_\phi,z_\psi)_i}(s,a)]$
    
        $\tilde{Z}(a) = \{(z_\phi,z_\psi) | (z_\phi,z_\psi) \in Z, a = \argmax\limits_{a^\star \in \mc{A}} \hat{Q}^\star_{(z_\phi,z_\psi)}(s,a^\star)\}$
        
        $\hat{v}_a(s) \triangleq \sum\limits_{a^\star \in \mc{A}} \frac{|\tilde{Z}(a^\star)|}{|Z|}(\frac{1}{|\tilde{Z}(a^\star)|} \sum\limits_{(z_\phi,z_\psi) \in \tilde{Z}(a^\star)} \hat{Q}^\star_{(z_\phi,z_\psi)}(s,a) - \frac{1}{|Z|} \sum\limits_{(z_\phi,z_\psi) \in Z} \hat{Q}^\star_{(z_\phi,z_\psi)}(s,a))^2$
    
        $\pi(\cdot | s_h) = \min\limits_{\pi \in \Delta(\mc{A})} \frac{\bE_{a \sim \pi}[(\hat{\Delta}_a(s))^2]}{\bE_{a \sim \pi}[\hat{v}_a(s)]}$
    
        Take action $a_h \sim \pi(\cdot|s_h)$ and observe $r_h, s_{h+1}$
    
        Sample random perturbations $\eta_\phi \sim \mc{N}(0, \sigma^2_\phi I), \eta_\psi \sim \mc{N}(0, \sigma^2_\psi I)$
    
        $\mc{D} \rightarrow \mc{D} \cup \{(s_h,a_h,r_h,s_{h+1},\eta_\phi, \eta_\psi)\}$
        
        Sample random minibatch $\tilde{\mc{D}} \stackrel{m}{\sim} \text{Uniform}(\mc{D})$
        
        
        
        $\nu \leftarrow \nu - \alpha \nabla_\nu \mc{L}(\nu,\tilde{\mc{D}})$
    }
}
\caption{Posterior State-Abstraction Sampling with Hypermodels \& Variance-IDS}
\label{alg:qmf}
\end{algorithm}

\end{document}